\documentclass[letterpaper, 10 pt, journal, twoside]{IEEEtran} \IEEEoverridecommandlockouts
\usepackage{cite}
\usepackage{amsmath,amssymb,amsfonts}
\usepackage{graphicx}
\usepackage{textcomp}
\usepackage{xcolor}
\usepackage{lipsum}
\usepackage{hyperref}
\usepackage{stfloats}
\usepackage{subcaption}
\usepackage{url}
\usepackage{verbatim}
\usepackage{layouts}
\usepackage{adjustbox}
\usepackage{makecell}
\usepackage{float}
\usepackage{xspace}
\usepackage{multirow}
\usepackage{pbox}
\usepackage{colortbl}
\usepackage{multicol}

\usepackage{booktabs}
\usepackage{tabularx}
\usepackage[absolute,overlay]{textpos}
\def\tablename{Table}

\usepackage[capitalize]{cleveref}

\usepackage{comment}
\usepackage[binary-units]{siunitx}
\usepackage{relsize}
\usepackage{ifthen}
\usepackage[colorinlistoftodos]{todonotes}

\usepackage[vlined,ruled,linesnumbered]{algorithm2e}
\usepackage{graphics} 
\usepackage{rotating}
\usepackage{color}
\usepackage{enumerate}
\usepackage[T1]{fontenc}
\usepackage{psfrag}
\usepackage{epsfig} 
\usepackage{booktabs}
\usepackage{graphicx,url}
\usepackage{array}
\usepackage{latexsym}
\usepackage{amsfonts}
\usepackage{amsmath}
\usepackage{amssymb}
\usepackage{xstring}
\usepackage[noend]{algorithmic}
\usepackage{multirow}
\usepackage{xcolor}
\usepackage{prettyref}
\usepackage{flexisym}
\usepackage{bigdelim}
\usepackage{breqn} 
\usepackage{listings}

\usepackage{enumitem}
\usepackage{xspace}
\usepackage{bm}
\graphicspath{{./figures/}}
\usepackage{tikz}
\usetikzlibrary{matrix,calc}

\usepackage{mdwlist}

\let\itemize\undefined
\makecompactlist{itemize}{stditemize}

\newrefformat{prob}{Problem\,\ref{#1}}
\newrefformat{def}{Definition\,\ref{#1}}
\newrefformat{sec}{Section\,\ref{#1}}
\newrefformat{sub}{Section\,\ref{#1}}
\newrefformat{prop}{Proposition\,\ref{#1}}
\newrefformat{app}{Appendix\,\ref{#1}}
\newrefformat{alg}{Algorithm\,\ref{#1}}
\newrefformat{cor}{Corollary\,\ref{#1}}
\newrefformat{thm}{Theorem\,\ref{#1}}
\newrefformat{lem}{Lemma\,\ref{#1}}
\newrefformat{fig}{Fig.\,\ref{#1}}
\newrefformat{tab}{Table\,\ref{#1}}

\newcommand{\bdmath}{\begin{dmath}}
\newcommand{\edmath}{\end{dmath}}
\newcommand{\beq}{\begin{equation}}
\newcommand{\eeq}{\end{equation}}
\newcommand{\bdm}{\begin{displaymath}}
\newcommand{\edm}{\end{displaymath}}
\newcommand{\bea}{\begin{eqnarray}}
\newcommand{\eea}{\end{eqnarray}}
\newcommand{\beal}{\beq \begin{array}{ll}}
\newcommand{\eeal}{\end{array} \eeq}
\newcommand{\beas}{\begin{eqnarray*}}
\newcommand{\eeas}{\end{eqnarray*}}
\newcommand{\ba}{\begin{array}}
\newcommand{\ea}{\end{array}}
\newcommand{\bit}{\begin{itemize}}
\newcommand{\eit}{\end{itemize}}
\newcommand{\ben}{\begin{enumerate}}
\newcommand{\een}{\end{enumerate}}

\newcommand{\etal}{\emph{et~al.}\xspace}

\newcommand{\eg}{\emph{e.g.,}\xspace}
\newcommand{\ie}{\emph{i.e.,}\xspace}

\newcommand{\hide}[1]{}

\newcommand{\hiddenText}{{\color{gray} hidden text.}}
\newcommand{\hideWithText}[1]{\hiddenText}

\newcommand{\blue}[1]{{\color{blue}#1}}

\newcommand{\linkToPdf}[1]{\href{#1}{\blue{(pdf)}}}
\newcommand{\linkToPpt}[1]{\href{#1}{\blue{(ppt)}}}
\newcommand{\linkToCode}[1]{\href{#1}{\blue{(code)}}}
\newcommand{\linkToWeb}[1]{\href{#1}{\blue{(web)}}}
\newcommand{\linkToVideo}[1]{\href{#1}{\blue{(video)}}}
\newcommand{\linkToMedia}[1]{\href{#1}{\blue{(media)}}}
\newcommand{\award}[1]{\xspace} 

\lstdefinelanguage{mine} 
{morekeywords={while,True,if,break,=,return,function,for,until,in,input,output,assumptions,assumption,invariant,loop,variant,end,invariant,precondition,variables}, 
sensitive=false, 
morecomment=[l]{\#}, 
morecomment=[il]\%{.}, 
morecomment=[s]{/*}{*/}, 
morestring=[b]", 
} 

\usepackage{calc}
\newlength\listingnumberwidth
\newif\ifarxiv
\arxivtrue

\ifarxiv
\else
\fi

\def\BibTeX{{\rm B\kern-.05em{\sc i\kern-.025em b}\kern-.08em
    T\kern-.1667em\lower.7ex\hbox{E}\kern-.125emX}}

\newcommand{\p}{\mathbb{P}}

\newcommand{\review}[1]{#1}
\newcommand{\final}[1]{#1}

\begin{document}
\ifarxiv




\begin{minipage}{\textwidth} 
\thispagestyle{empty}
\copyright 2024 IEEE. Personal use of this material is permitted. Permission from IEEE must be obtained for all other uses, in any current or future media, including reprinting/republishing this material for advertising or promotional purposes, creating new collective works, for resale or redistribution to servers or lists, or reuse of any copyrighted component of this work in other works.\\ Please cite this paper as:\\
\begin{verbatim} 
@ARTICLE{Gorlo2024LP2,
  author={Gorlo, Nicolas and Schmid, Lukas and Carlone, Luca},
  journal={IEEE Robotics and Automation Letters}, 
  title={Long-Term Human Trajectory Prediction Using 3D Dynamic Scene Graphs}, 
  year={2024},
  volume={9},
  number={12},
  pages={10978-10985},
  doi={10.1109/LRA.2024.3482169}
  }
\end{verbatim} 
\end{minipage}
\newpage

\fi

\title{
Long-Term Human Trajectory Prediction\\using 3D Dynamic Scene Graphs
}

\author{Nicolas Gorlo, Lukas Schmid, and Luca Carlone%

\thanks{The authors are with the MIT SPARK Lab, Massachusetts Institute of Technology, Cambridge, USA. {\tt \footnotesize \{ngorlo,lschmid,lcarlone\}@mit.edu}}%
\thanks{This work was partially supported by Amazon, Lockheed Martin, and the Swiss National Science Foundation (SNSF) grant No. 214489.}%

}

\maketitle

\ifarxiv
{
\begin{textblock*}{\textwidth - 10mm}(\marginparwidth,5mm) 
\noindent
\small \copyright 2024 IEEE. Personal use of this material is permitted. Permission from IEEE must be obtained for all other uses, in any current or future media, including reprinting/republishing this material for advertising or promotional purposes, creating new collective works, for resale or redistribution to servers or lists, or reuse of any copyrighted component of this work in other works.
\end{textblock*}
\pagenumbering{arabic}}
\fi

\begin{abstract}
    
We present a novel approach for long-term human trajectory prediction \review{in indoor human-centric environments}, which is essential for long-horizon robot planning in \review{these} environments. 
State-of-the-art human trajectory prediction methods are limited by their focus on collision avoidance and short-term planning, and their inability to model complex interactions of humans with the environment. 
In contrast, our approach overcomes these limitations by predicting sequences of human interactions with the environment and using this information to guide trajectory predictions over a horizon of up to \SI{60}{\second}. 
We leverage Large Language Models (LLMs) to predict interactions with the environment by conditioning the LLM prediction on rich contextual information about the scene.
This information is given as a 3D Dynamic Scene Graph that encodes the geometry, semantics, and traversability of the environment into a hierarchical representation.
We then ground these interaction sequences into multi-modal spatio-temporal distributions over human positions using a probabilistic approach based on continuous-time Markov Chains.
To evaluate our approach, we introduce a new semi-synthetic dataset of long-term human trajectories in complex indoor environments, which also includes annotations of human-object interactions.
We show in thorough experimental evaluations that our approach achieves a 54\% lower average negative log-likelihood and a 26.5\% lower Best-of-20 displacement error compared to the best non-privileged \review{(i.e., evaluated in a zero-shot fashion on the dataset)} baselines for a time horizon of \SI{60}{s}.
\end{abstract}

\section{Introduction}\label{sec:introduction}
\IEEEPARstart{T}{he} ability to predict human motion in complex human-centric environments is a crucial component for scene understanding and autonomy, and is prerequisite to a number of applications ranging from consumer, assistive, and care robotics to industrial and retail robots or augmented and virtual reality.
Most notably, fulfilling tasks proactively and operating efficiently in human-populated environments, like intercepting a human or moving out of the way proactively, requires reliable predictions of human trajectories over a long time-horizon.

Most current works on human motion and trajectory prediction are primarily motivated by collision avoidance and short-term planning and hence focus on predicting trajectories up to a maximum of $\SI{5}{s}$~\cite{Rudenko20ijrr-humanTrajectoryPrediction,Cao20eccv-longterm,Salzmann23ral-hst,Sadeghian19cvpr-SoPhie,Yuan21iccv-AgentFormer, Salzmann20eccv-trajectronpp}.
Thus, trajectories in current benchmark datasets are \review{oftentimes not} longer than $\SI{10}{s}$~\cite{Pellegrini09iccv-ethdataset, Lerner07wiley-ucydataset}.
\review{Only recently and rarely, longer prediction horizons up to \review{$\SI{60}{s}$} have been explored~\cite{Rudenko20ijrr-humanTrajectoryPrediction, mangalam21iccv-ynet}.
However, such predictions are highly beneficial to effective long-term planning and proactive long-term behavior of robots in complex social environments. }

Further, most \review{current} approaches focus on environments with primarily smooth trajectories, such as outdoor scenes~\cite{robicquet16eccv-stanfordDrones}, \review{or large-scale indoor scenarios, where the structure of the environment (e.g., walkways and hallways) primarily influences the flow of the trajectories rather than interactions of humans with objects in the environment~\cite{Zhou12cvpr-CentralStationDataset, brvsvcic13thms-ATCShoppingDataset}}. 
In contrast, human-centric environments are oftentimes geometrically complex and semantically rich, leading to more complex trajectories \review{and interactions with the scene}.

\begin{figure}[]
\centering
\includegraphics[width=.95\linewidth]{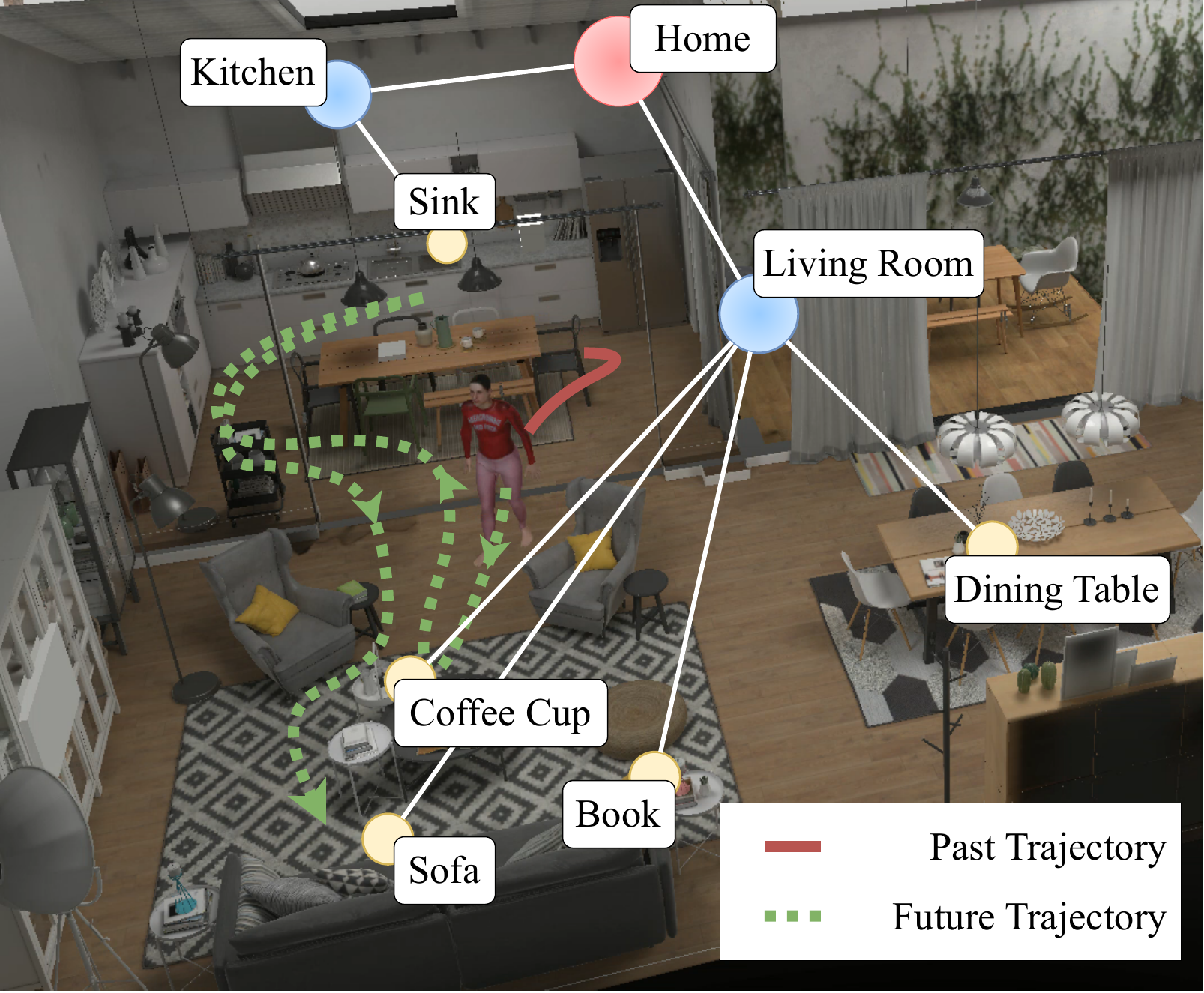} 
\caption{Our method, \emph{LP$^2$}, predicts a spatio-temporal distribution over long-term (up to $\SI{60}{s}$) human trajectories in complex environments by reasoning about their interactions with the scene, represented as a 3D Dynamic Scene Graph.}\label{fig:titlefig}
\vspace{-10px}
\end{figure}

Predicting trajectories $\SI{60}{s}$ into the future is a challenging task.
First, the aleatoric uncertainty quickly increases with time.
Second, the distribution over plausible or even likely human trajectories is highly multi-modal.
In addition, humans are much more likely to engage in complex interactions with their environments during longer time horizons which is not well captured by most existing methods.
For example, it is feasible to predict human motion if the human follows a corridor, but most methods fail when the human stops to interact with an object in the corridor (e.g., a water dispenser).

To address these challenges, we propose a novel \review{zero-shot} approach to long-term human trajectory prediction called \emph{``Language-driven Probabilistic Long-term Prediction'' (LP$^2$)} that takes such environmental factors into consideration.
In contrast to directly predicting human trajectories from past trajectories, we reason about sequences of human interactions with the environment.
To this end, we leverage recent developments in 3D dynamic scene graphs (DSGs) \cite{Armeni19iccv-3DsceneGraphs,Rosinol21ijrr-Kimera,Hughes24ijrr-hydraFoundations, Looper23icra-variableSceneGraph} as a scene representation capturing symbolic concepts such as traversability, objects, and agents, as well as advances in Large Language Models (LLMs) for reasoning.
LLMs have shown promise in modeling human behavior~\cite{Park23uist-GenerativeAgents} and anticipating long-term actions~\cite{Kim25eccv-palm,Kim23arxiv-lalm}. \review{Very recently, LLMs have garnered interest in the area of human trajectory prediction. Concurrent to our work, Bae~\etal~\cite{Bae24cvpr-LMTraj} present a trajectory prediction framework that transforms trajectory coordinates and scene images into textual prompts, while Chib~\etal~\cite{Chib24arxiv-LGTraj} use LLMs to generate motion cues used in a transformer-based architecture. These methods, however, are predicting trajectories for a time horizon of \SI{4.8}{s}. Moreover, Lan~\etal~\cite{Lan24itiv-TrajLLM} present a novel approach for vehicle trajectory prediction without explicit prompt-engineering of the LLM. In contrast, we} leverage LLMs in a novel auto-regressive scheme to predict likely interactions with the scene, and propose an approach based on continuous-time Markov chains (CTMC) to \review{then} infer a probability distribution of the human location across space and time.
We further create a novel semi-synthetic dataset of long-term human trajectories in complex indoor environments with rich scene and interaction annotations. 
We show in thorough experimental evaluations that our approach achieves a $54\%$ lower average negative log-likelihood (NLL) and a $26.5\%$ lower Best-of-20 displacement error compared to the best non-privileged \review{(i.e., zero-shot on the dataset)} baselines for a time horizon of \SI{60}{s}.

In summary, we make the following contributions:
For the first time, we study long-term human trajectory prediction \review{in complex, human-centric indoor environments} with a prediction horizon of up to $\SI{60}{s}$, including complex human-object interactions. We present \emph{LP$^2$}, a novel method combining LLM-based zero-shot interaction sequence and CTMC-based probabilistic trajectory prediction to infer multi-modal spatio-temporal distribution over future human positions. \review{Finally,} \final{w}e introduce a new semi-synthetic dataset of human trajectories in complex indoor environments with annotated human-object interactions. We release our method and data 
open-souce\footnote{Released at \url{https://github.com/MIT-SPARK/LP2} upon acceptance.}.

\section{Related Works}

\textbf{Long-term Human Trajectory Prediction.}
The problem of long-term human trajectory prediction has attracted notable research interest, and various techniques, ranging from physics-based to pattern-based and planning-based approaches, have been proposed~\cite{Rudenko20ijrr-humanTrajectoryPrediction}.
However, the term "long-term" is oftentimes used to describe trajectories or joint-level motion of a length of \SI{5}{s} or less~\cite{Salzmann20eccv-trajectronpp, Cao20eccv-longterm} and sometimes up to \SI{12}{s}~\cite{Tran21wacv-goal}. \review{Mangalam~\etal~\cite{mangalam21iccv-ynet}, present results on long-term prediction up to \SI{60}{s}. However, their predictions are limited to monotonous environments like corridoors or outdoor trails and may fail in more complex environments.
In contrast, we address the challenge of long-term (up to \SI{60}{s}) trajectory prediction in complex human-centric indoor environment.}

Recent advances focus on integrating additional input modes and other priors \ie ``target agent cues''~\cite{Rudenko20ijrr-humanTrajectoryPrediction}, like eye-gaze, human joint-level poses~\cite{Salzmann23ral-hst}, or scene context~\cite{Cao20eccv-longterm,Salzmann23ral-hst,Sadeghian19cvpr-SoPhie,Salzmann20eccv-trajectronpp, mangalam21iccv-ynet} expressed as occupancy maps~\cite{Salzmann20eccv-trajectronpp, mangalam21iccv-ynet, DeBrito21corl-socialvrnn}, images~\cite{Sadeghian19cvpr-SoPhie, mangalam21iccv-ynet, Salzmann20eccv-trajectronpp} or LiDAR scans~\cite{Salzmann23ral-hst}. \review{Further, \final{De Almeida}~\etal~\cite{DeAlmeida23iccv-thormagni} leverage knowledge of human roles like ``carrier'' or ``visitor'' to better predict human trajectories in indoor environments, while~\cite{Narayanan20arxiv-emotion} explores using information about the target agents state of affect and intent.}
A group of recent works addresses the problem of social human trajectory forecasting, achieving improved results by modeling the influences of other humans on human trajectories~\cite{Gupta18cvpr-social, Sadeghian19cvpr-SoPhie, Salzmann20eccv-trajectronpp, Yuan21iccv-AgentFormer}. 
However, they typically still focus on a comparably short prediction horizons of 3-\SI{20}{s} and smooth unidirectional motion. 
\review{Alternatively, a set of methods~\cite{Zhu23iros-clifflhmp, Zhu24icra-lace} utilizes previously observed trajectories of humans in the target scene to infer common motion patterns and predict at a long horizon.
While this can be helpful information to improve trajectory prediction, these approaches cannot be easily transferred to zero-shot prediction in novel scenes.} 
In contrast, we address the problem of \review{zero-shot} long-horizon (up to \SI{60}{s}) trajectory prediction, where complex human-object interactions may lead to sudden stops or changes in walking direction \review{and propose using past interactions with the environment as cues for trajectory prediction.
Predicting human navigation goals via human-object interactions has been explored by Bruckschen~\etal~\cite{Bruckschen20ras-humannavigationgoals}, proposing a Bayesian inference framework to predict the goal of an agent based on previously observed human-object interactions in the environment.
However, this method relies on previously observed interactions with the same objects.}

\textbf{Datasets for Human Trajectory Prediction.}
While there are many outstanding datasets for human trajectory prediction\cite{brvsvcic13thms-ATCShoppingDataset,robicquet16eccv-stanfordDrones, Pellegrini09iccv-ethdataset, Lerner07wiley-ucydataset,Rudenko20ral-thor,Schreiter24ijrr-thormagni,Finean22ras-oxfordihm,Hassan19iccv,Yan17iros-lcas,Martin21pami-jrdb,Karnan22ral-scand,Hassan21iccv-samp,Cao20eccv-longterm}, the presented problem requires a number of specific attributes: trajectories need to extend over long time spans ($\geq\SI{60}{s}$) and take place in complex and semantically rich human-centric environments, such as an office or a residential space.
In addition, sensory data that allows the construction of a scene representation and rich annotations such as ground truth scenes, trajectories, and interactions should be provided.

Several datasets provide real-world human trajectories, but lack enough sensory data about the environment~\cite{Pellegrini09iccv-ethdataset, Lerner07wiley-ucydataset} and are often relatively short term. 
Other datasets cover longer time horizons and interactions with the environment, but are restricted to single-room, lab-like environments~\cite{Rudenko20ral-thor,Schreiter24ijrr-thormagni,Finean22ras-oxfordihm,Hassan19iccv}. 
\review{A third category covers large indoor areas (like a shopping mall) or widespread outdoor areas\cite{brvsvcic13thms-ATCShoppingDataset,robicquet16eccv-stanfordDrones}. Evaluating on these datasets cannot generalize to complex human-centric indoor environments, where the target agents frequently interact with entities in their environments and change the heading of their trajectory as a result.}
To achieve detailed annotations, synthetic datasets have been recorded~\cite{Hassan21iccv-samp,Cao20eccv-longterm}, but mostly focus on joint-level trajectory prediction.
In contrast, this work introduces a novel semi-synthetic dataset of 76 long-term trajectories (\SI{3}{\minute}) in two semantically rich environments.

\section{Problem Statement}\label{sec:problem_formulation}

We address the task of long-term human trajectory prediction in semantically rich environments, 
given the human's past trajectory, past interactions with the environment, and a model of that environment.
We denote the future trajectory $\mathcal{T}_f = \{\mathbf{y}_{t+1},\mathbf{y}_{t+2},\ldots,\mathbf{y}_{t+T}\}$ and past trajectory $\mathcal{T}_p = \{\mathbf{y}_{t-T_0}, \mathbf{y}_{t-T_0+1}, \ldots, \mathbf{y}_{t}\}$, where $\mathbf{y}_i \in \mathbb{R}^2$ denotes the 2D position of the human at time $i$.
We further assume that past human-object interactions $A_p$ (encoded by the associated object), action description, and duration triples (e.g., \{"sink", "wash hands", \SI{12}{s}\}), and a scene representation $S$ is given.
Since many future trajectories are possible, the goal is to estimate a multi-modal spatio-temporal distribution $\p(\mathcal{T}_f | A_p, \mathcal{T}_p, S)$ over future trajectories $\mathcal{T}_{f}$ given the past interactions $A_p$, past trajectory $\mathcal{T}_p$, and scene representation $S$.
%
%
We assume there is a single human in the environment.\footnote{While outside the scope of this work, we note that dynamic entities, such as other agents, could easily be included in the scene representation $S$.}
\section{Approach}
\label{sec:approach}

Our goal is to predict long-term future human trajectories, where humans may interact with several objects in the scene leading to sudden stopping or changes in moving direction.
As there are many plausible objects to interact with and varying durations of these interactions, the prediction of a highly multi-modal distribution both in space and time is required.

\begin{figure}
    \centering
    \includegraphics[width=\linewidth]{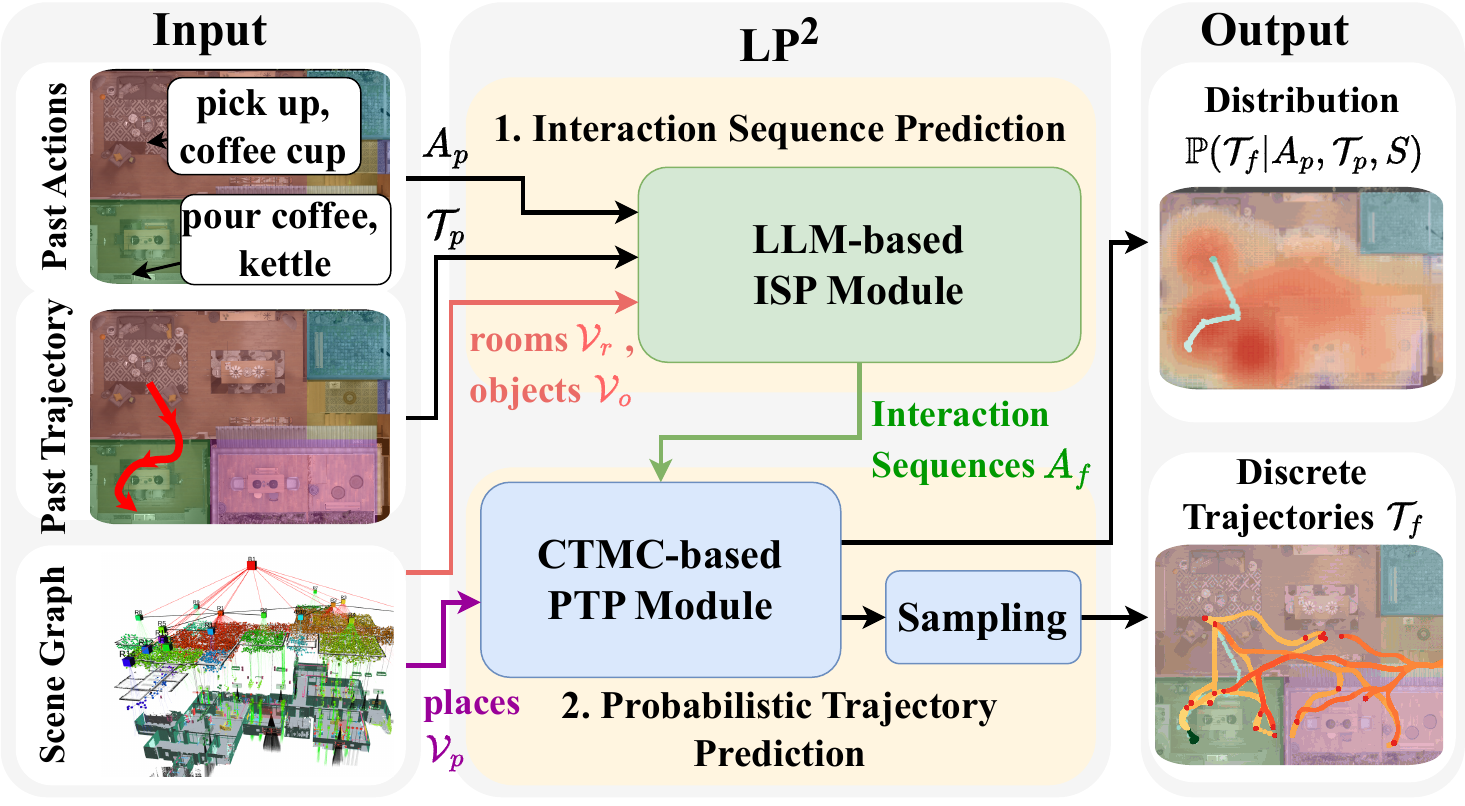}
    \caption{\review{Approach overview.} The interaction sequence prediction (ISP) module estimates sequences of future interactions with the environment using the rich semantic information of the scene graph $S$. The probabilistic trajectory prediction (PTP) module connects these sequences into cohesive trajectories and predicts a continuous spatio-temporal probability distribution over the future human position in the environment.}
    \label{fig:high_level_architecture}
    \vspace{-10px}
\end{figure}

To overcome these challenges, we propose a hierarchical approach that splits the task into high- and low-level reasoning parts.
First, an \emph{interaction sequence prediction} (ISP) module predicts sequences of likely future interactions $A_f$ of the human in the environment.
Second, a \emph{probabilistic trajectory prediction} (PTP) module grounds these sequences into cohesive trajectories $\mathcal{T}_f$ and predicts a continuous spatio-temporal probability distribution $\p(\mathcal{T}_f | A_f, A_p, \mathcal{T}_p, S)$ over the future human position. 
Formally, this reflects the assumption that humans are goal driven, \ie their trajectories are fully specified by their environment and goals, where goals in turn can be inferred from their previous activities:
\begin{equation}
    \p(\mathcal{T}_f | A_p, \mathcal{T}_p, S) = \sum_{A_f} \underbrace{\p(\mathcal{T}_f | A_f, \mathcal{T}_p, S)}_{\text{PTP}} \underbrace{\p(A_f | A_p, \mathcal{T}_p, S)}_{\text{ISP}} \label{eq:decomposition}    
\end{equation}
An overview of our approach is shown in Fig.~\ref{fig:high_level_architecture}.

\textbf{Environment Representation.}
To effectively reason about complex scenes we choose 3D dynamic scene graphs (DSGs)~\cite{Armeni19iccv-3DsceneGraphs,Rosinol21ijrr-Kimera} as a scene representation.
A DSG provides a hierarchical symbolic abstraction, where each symbol is physically grounded in the scene.
Formally, it consists of a set of nodes and edges $S = \{\mathcal{V}, \mathcal{E}\}$.
The nodes $v\in \mathcal{V}$ are grouped into layers $L$, with inter- and intra-layer edges $e_{ij} \in \mathcal{E}$ specifying relationships between nodes $v_i$ and $v_j$.

We follow the definition of \cite{Hughes24ijrr-hydraFoundations} with layers $L=\{ o,p,r\}$ encoding objects $o$, places $p$, and rooms $r$, respectively. 
Notably, all nodes $v_o \in \mathcal{V}_o$ denote a semantic object in the scene, whereas all places $v_p\in \mathcal{V}_p$ and their intra-layer edges $e_{p_ip_j} \in \mathcal{E}$ represent the traversable free space in the scene.

While we assume the DSG $S$ to be given in this work, it can also be constructed in real-time on a robot~\cite{Hughes22rss-hydra,Hughes24ijrr-hydraFoundations}.

\begin{figure}
    \centering
    \includegraphics[width=.9\linewidth]{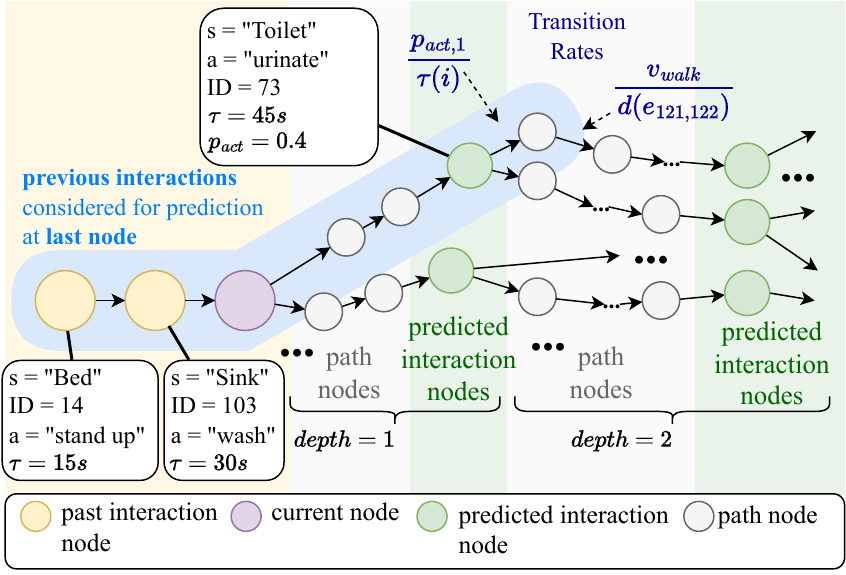}
    \caption{\review{Interaction sequence tree $T_I$.}
    The LLM is auto-regressively prompted with the scene $S$ and the previous interactions $A_p$ to predict the next interactions (green nodes). 
    Shortest paths spatially connect interactions (gray nodes). 
    Progression through the tree over time is modeled as a CTMC with probability-based transition rates (blue). 
    As time progresses, the probability mass moves from left to right. 
    }
    \label{fig:interaction_tree}
    \vspace{-10px}
\end{figure}

\textbf{Interaction Sequence Prediction.}
~\label{sec:interaction_sequence_prediction_module}
To estimate future interactions, we build on the grounded symbols in the DSG $S$ and the zero-shot reasoning power of LLMs~\cite{Park23uist-GenerativeAgents,Kim25eccv-palm} for human action forecasting.
Similar to Rana~\etal~\cite{Rana23corl-sayPlan}, we first parse $S$ into a natural language description of the scene $\widetilde{S}$. In contrast to \cite{Rana23corl-sayPlan}, we write full sentence descriptions of the environment, hierarchically describing all rooms $v_r \in \mathcal{V}_r$, connectivity between rooms $e_{r_ir_j} \in \mathcal{E}$, and objects $v_o \in \mathcal{V}_o\ |\ \exists e_{v_ov_r} \in \mathcal{E}$ in each room. 
\review{An example of $\widetilde{S}$ then reads: {\small{``\texttt{In the environment, there are the rooms: <rooms>. room connections: <room connections>. These objects are in the environment: <objects per room>}.''}}}
We prompt the LLM to predict $W_I \in \mathbb{N}^+$ next interactions $A_i$ and likelihoods $p_{act, i} \in \left(0,1\right]$ of $A_i$ occurring as well as estimated durations $\tau_i$ of $A_i$ based on the description of the environment $\widetilde{S}$ and the past interactions $A_p$. 
We further instruct the LLM to provide reasoning about its prediction, which we found improves the prediction quality. 
The output is a JSON serializable dictionary of $A_i$, an example of which is shown as a green node in~\cref{fig:interaction_tree}.

We consider two levels of granularity when predicting the objects of $A_i$, prompting the LLM to predict either only the \emph{semantic} class $s$ of the next object or the unique object \emph{instance}.
In the former case, we conjecture that humans are more likely to interact with close-by entities and consider the $N_{s}$ closest instances of the predicted class $s$ as target objects, where we additionally weight the predicted likelihood $p_{act, i}$ by the inverse distance. 
To simplify the LLM prediction task, we also condense the scene description $\widetilde{S}$ in this case for objects of each semantic class in a room into a single description, \eg ``\texttt{In the kitchen, there are 4 chairs}''.

To predict a sequence of interactions, we build a tree $T_I$ with the agent's current position as a root and all predicted next interactions as leaves.
For each leaf in $T_I$, we auto-regressively prompt the LLM with the accumulated previous actions of the human to predict the following interactions, up to a \emph{depth} of $D_I$. 
This results in up to $(W_I \times N_s)^{D_I}$ total considered interaction sequences.

\textbf{Probabilistic Trajectory Prediction.}
Finally, we need to translate the interaction tree $T_I$ into a spatio-temporal distribution of human positions.
To ground the interaction predictions in spatial positions, we augment $T_I$ with the traversability information captured in the places $\mathcal{V}_p$ of $S$.
Assuming humans are \review{locally} efficient in their paths, we find the shortest-path in $\mathcal{E}_{pp}$ connecting each previous interaction to the possible following ones.
A resulting tree is shown in Fig.~\ref{fig:interaction_tree}, where each green node corresponds to a future interaction $A_i$, and gray nodes represent the traversable paths connecting them.

We model the human's progression through this tree over time as a stochastic process and construct a continuous-time Markov chain (CTMC)~\cite{norris98wiley-markov}. 
The CTMC is defined by the state space $X$, given by the nodes of the interaction tree $T_I$, an initial state~$x_0$, and a transition matrix~$Q \in \mathbb{R}^{|X|\times|X|}$.
Initially, all probability mass is concentrated at the current position of the human $\p(x(0)) = \{1, 0^{|X|-1}\}$.
We model the time to traverse a path as an exponential distribution where the mean is the distance divided by the average human walking speed $v_{walk}$.
Similarly, we model the duration of an interaction $A_i$ as an exponential distribution based on the predicted duration $\tau_i$.
The transition matrix can thus be constructed as:
\begin{align}
    Q_{i,j} =
    &~\begin{cases}
    \frac{v_{walk}}{d(x_i, x_j)} & \text{if }x_i \text{ \& }x_j \text{ are path nodes} \\
    \frac{p_{act, j}}{\tau_i} & \text{if } i \text{ is an interaction node}\\
    \sum\limits_{x_{i} \in X,~i\neq j}-Q_{i,j} & \text{if } i = j \\
    0 & \text{else }
    \end{cases},
\end{align}

\noindent
where $d(x_i, x_j)$ is the Euclidean distance between nodes $x_{i}$ and $x_{j}$.
To infer the probability distribution $\p(x(t))$ of the human being at state $x$ at time $t$, we solve the Kolmogorov forward equation~\cite{norris98wiley-markov} given by
\begin{equation}
    \frac{\partial}{\partial t} \p(x(t)) = Q\cdot \p(x(t)).
\end{equation}
The solution to this equation is
\begin{equation}
    \p(x(t)) = \exp(Q(t-t_0)) \p(x_0).
\end{equation}
As our structure of the CTMC is a tree, the transition matrix $Q$ is triangular and the exponential can be computed efficiently.

\begin{figure}
    \centering
    \includegraphics[width=\linewidth]{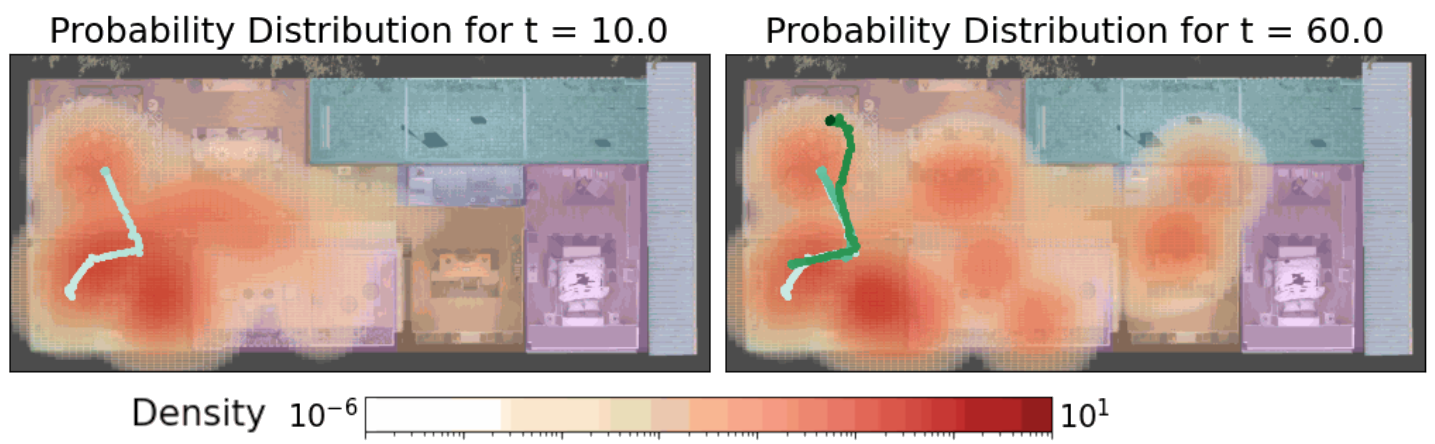}
    \caption{Spatio-temporal distribution over the human's predicted position (orange) progressing through time. True trajectory colored from light at $t=\SI{0}{s}$ to dark at $t=\SI{60}{s}$ green.
}\label{fig:output_distribution}
\vspace{-10px}
\end{figure}

To also consider spatial uncertainty, we model the probability of the human being in physical space $\xi$ at time $t$ as a mixture of Gaussians with positions $x$, weights $\p(x(t))$, and an empirically determined kernel width.
An example of the resulting spatio-temporal distribution is shown in Fig.~\ref{fig:output_distribution}.

\textbf{Implementation Details.}
We use GPT-4-turbo~\cite{OpenAI23arxiv-gpt4} as a state-of-the-art LLM in our ISP module.
The configurable ISP hyperparameters are set to $W_I=6$, $D_I=2$, and $N_s=3$, resulting in up to 324 considered future interaction sequences.

\section{Dataset}
As there currently exist no datasets for our task, a novel semi-synthetic dataset is created.
We consider two typical environments: a large office and a smaller home scene.
They span 13 rooms with 107 objects over $\SI{994}{m^2}$ and 9 rooms with 120 objects over $\SI{111}{m^2}$, respectively.
To create realistic and diverse trajectories, we generate interaction sequences by human annotators, crowd-sourced through Amazon Mechanical Turk.
To achieve coherent and realistic trajectories, annotators are prompted to envision sequences of interactions with objects in a pictured scene while fulfilling an abstract task in that scene. 
Interactions are annotated with estimates of interaction times, action descriptions, and a description of the human's higher-level goal.
We manually verify the quality of all submissions.
We then simulate SMPL human shape models~\cite{Loper15tg-smpl} fulfilling these sequences of interactions in TESSE~\cite{ravichandran20arxiv-tesse}, a high-fidelity simulator based on unity.

\begin{figure}
  \centering
  \includegraphics[width=.95\linewidth]{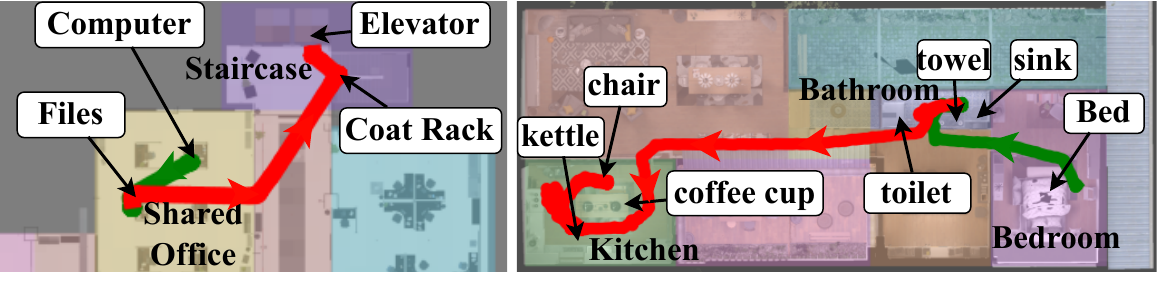}
  \caption{Example trajectories in the office (left) and home (right). Trajectory is shown in green (past) and red (future).}
  \label{fig:dataset_samples}
\end{figure}

Our dataset covers a wide range of scenarios, including trajectories that start while the human is interacting as well as when the human is moving in between interactions.
To prevent bias towards the start point, we split trajectories into past/future after the second interaction or \SI{60}{s} before the end of the full recorded trajectory, whichever comes first. This way, we focus on the part of the trajectory that is relevant to most applications. 
The dataset contains a total of 76 trajectories, whose diverse characteristics are shown in Tab.~\ref{tab:dataset_properties}. 
\review{To demonstrate the diverse distribution that arose from the data collection effort, we evaluate the normalized Levenshtein distance, \ie the fraction of the interaction sequence that needs to be modified to arrive at the next closest sample, and the path efficiency \cite{Schreiter24ijrr-thormagni}. Dataset metrics and example trajectories are shown in Tab.~ \ref{tab:dataset_properties} and Fig.~\ref{fig:dataset_samples}, respectively.}
\final{While 76 trajectories may be insufficient to train large models from scratch, we show in \cref{sec:experiments} that our dataset can be used to evaluate zero-shot methods and fine-tune pre-trained models.}




\begin{table}
\adjustbox{width=\columnwidth}{%
\begin{tabular}{lcc}
    Attribute & Office & Home \\ \midrule
    \#Trajectories that start while interacting [1] & 11 & 14 \\
    \#Trajectories that start while walking [1] & 34 & 17 \\
    Distance travelled in past [m] & 14.0$\pm$11.9 & 7.8$\pm$3.0 \\
    Distance travelled in future [m] & 18.6$\pm$12.7 & 11.7$\pm$6.5 \\
    \#Past interactions [1] & 2.2$\pm$0.5 & 2.4$\pm$0.5 \\
    \#Future interactions [1] & 2.3$\pm$0.8 & 2.2$\pm$0.8 \\
    Duration of past observation [s] & 67.6$\pm$ 30.8 & 60.6$\pm$30.2\\
    Time spent interacting in past [s] & 55.2$\pm$30.1 & 55.1$\pm$30.4 \\ 
    Time spent interacting in future [s] & 43.7$\pm$11.5 & 51.3$\pm$6.6\\
    \review{Average path efficiency \final{\cite{Schreiter24ijrr-thormagni, Amirian20accv-opentraj}}} & \review{0.182$\pm$0.112} & \review{0.116$\pm$0.073} \\
    \review{Average Levenshtein distance to closest sample} & \review{0.678} & \review{0.585} \\
\end{tabular}
}
\caption{Diverse characteristics of our newly recorded dataset.}\label{tab:dataset_properties}
\vspace{-10px}
\end{table}

\section{Experiments}
\label{sec:experiments}

\textbf{Experimental Setup.} 
In order to focus on the performance of the trajectory prediction, a DSG $S$ is constructed for each scene based on ground-truth semantic information provided by the simulator. 
Akin to relational action forecasting~\cite{Sun19cvpr-relational}, we separate the task of predicting future actions from action recognition and assume the past interactions to be given. 
We note that when engineering an end-to-end system, a DSG, past trajectories and interactions could be estimated using approaches like~\cite{Hughes24ijrr-hydraFoundations, Dang20pr, Feichtenhofer22neurips, Schmid24rss-khronos}, but engineering such a system is currently out of scope for this work.
Due to the different scales of the environments we evaluate each one separately.
\review{The length of $\mathcal{T}_p$ is given by the past actions of the human, with a minimum length of \SI{8.1}{s} in the office and \SI{10.75}{s} in the home environment.}

\textbf{Metrics.} 
We evaluate the ability to predict multi-modal spatio-temporal distributions by computing the \emph{negative log likelihood} (NLL) of the ground truth future trajectory in the predicted distribution. 
We further compute the \emph{best-of-N average displacement error} (BoN ADE), reporting the lowest mean squared error between of N discrete predicted trajectories and the ground truth.
Both metrics are computed over \review{different time horizons} and lower is better.

\textbf{Baselines.} 
Comparing our method to prior art in long-term human trajectory prediction is challenging, as most methods rely on training their models on extensive datasets. 
As we are not aware of other datasets that provide human trajectories over \SI{60}{\second}, we train these methods on existing datasets and fine tune them on our dataset.
We compare against \emph{Trajectron++}~\cite{Salzmann20eccv-trajectronpp} and \emph{YNet}~\cite{mangalam21iccv-ynet}, two state-of-the-art methods for human trajectory forecasting.
We train Trajectron++ on the ETH and UCY datasets~\cite{Pellegrini09iccv-ethdataset,Lerner07wiley-ucydataset}, including occupancy maps of the dataset scenes as scene input. 
Additionally, we fine tune the model on the hold-out scenes of our dataset. 
Since YNet's architecture requires a fixed prediction horizon, we train it on the Stanford drones dataset~\cite{robicquet16eccv-stanfordDrones} providing trajectories up to \review{\SI{60}{s}} into the future. 
We also include occupancy maps of the dataset scenes as input and fine tune YNet on our dataset for fair comparison. 
Note that our method has not seen any data of our dataset while both of the baselines can be considered privileged. 
We further compare to a \emph{constant velocity} model based on the current direction and velocity of the human, and also provide a \emph{random walk} and motion towards a \emph{random goal} within the places of the scene graph as references.
The NLL of these baselines is measured as for our method by performing Kernel Density Estimation (KDE) over the predicted trajectories for each timestep.

\begin{figure}
    \centering
    \includegraphics[width=\linewidth]{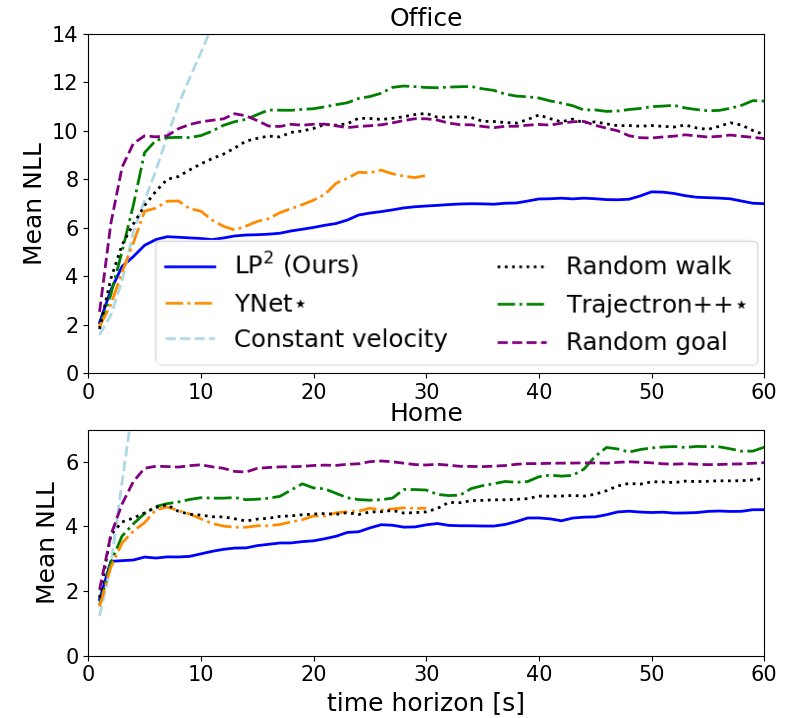}
    \caption{Negative log-likelihood evolving over time.}
    \label{fig:results_nll}
    \vspace{-10px}
\end{figure}

\textbf{1) Long-term Trajectory Prediction Performance.}
To investigate the ability to predict multi-modal spatio-temporal distributions over complex human trajectories, we show the average NLL vs. future time in each environment in Fig.~\ref{fig:results_nll}.
We observe that for short horizons of $\sim \SI{4}{s}$ most methods achieve similar performance.
The only outlier is \emph{random goal}, which in some cases may send the prediction off into a wrong direction.
However, as the prediction horizon increases our approach shows superior performance compared to all baselines.
Notably, the NLL only increases gradually for \emph{LP$^2$}, demonstrating its ability to support multi-modal long-horizon distributions.
In contrast, most other methods quickly saturate at an uninformative state as indicated by the \emph{random goal} baseline.
This can be explained by the fact that in our setting, where trajectories are less unidirectional and homogeneous, Trajectron++ and YNet struggle to account for sudden stops or changes in walking direction.
While it may seem counter-intuitive that the \emph{random walk} performs strongly, this represents an inherent bias in humans oftentimes not moving very far away from their starting point when accomplishing a task.
As expected, this is more pronounced for the smaller home environment.


We further show the Bo5 and Bo20 ADE at \review{various prediction time intervals} for methods that output discrete trajectories in Tab.~\ref{tab:results_ade}. 
We observe that \emph{LP$^2$} shows strong performance for ADE over the entire length of $\SI{60}{s}$, most pronounced for the multi-modal Bo20 setting.
The baseline methods show better performance early, which is the domain they specialize in.
While our method does not focus on taking the past trajectory much into account for accurate prediction of the immediate next trajectory, we believe that such approaches can be combined with our long-term predictions.
It is also worth pointing out that due to our tree-structured ISP, multiple of the most likely trajectories may have the same first interaction, reducing the breadth of our early prediction.
Finally, both YNet and Trajectron++ can achieve strong performance when fine tuned on the hold-out scenes of the dataset.
However, the performance is notably decreased without this additional domain knowledge, whereas \emph{LP$^2$} achieves accurate predictions in a zero-shot manner.

\setlength{\tabcolsep}{4pt} 
\begin{table}[]
\adjustbox{width=\columnwidth}{%
\begin{tabular}{p{0.1em}l|ccc|ccc}
    \toprule
    \multirow{2}{0.1em}{\rotatebox[origin=c]{90}{Scene}} & Metric & \multicolumn{3}{c}{Bo5 ADE [m]} & \multicolumn{3}{c}{Bo20 ADE [m]} \\
    \cmidrule(lr){3-5} \cmidrule(lr){6-8}
    & Timeframe [s]& 0-10 & 10-30 & 30-60 & 0-10 & 10-30 & 30-60\\
    \midrule
    \multirow{6}{0.1em}{\rotatebox[origin=c]{90}{Office}} 
    & Trajectron++$^\star$ & 2.60 & 5.30 & \textbf{6.94} & 2.55 & 4.93 & 6.25 \\
    & YNet$^\star$ & \textbf{2.34} & \textbf{4.64} & - & \textbf{1.92} & \textbf{2.35} & - \\
    \arrayrulecolor{lightgray} \cmidrule(l{.7em}){2-8} \arrayrulecolor{black}
    & Trajectron++ & 2.75 & 8.34 & 16.4 & \underline{2.42} & 5.90 & 10.9 \\
    & YNet & 3.41 & 9.63 & - & \final{3.05} & \final{8.07} & - \\
    & Const. Vel. & \underline{2.67} & 10.6 & 23.7 & 2.67 & 10.6 & 23.7\\
    & \emph{LP$^2$} (Ours) & 3.04 & \underline{6.38} & \underline{7.70} & 3.00 & \underline{4.43} & \textbf{4.06} \\
    \midrule
    \multirow{6}{0.1em}{\rotatebox[origin=c]{90}{Home}} 
    & Trajectron++$^\star$  & 1.39 & 1.59 & 1.99 & 1.33 & 1.34 & 1.67 \\
    & YNet$^\star$ & 1.40 & 2.04 & - & 1.33 & 1.30 & - \\
    \arrayrulecolor{lightgray} \cmidrule(l{.7em}){2-8} \arrayrulecolor{black}
    & Trajectron++ & 2.23 & 6.51 & 14.3 & 1.67 & 4.03 & 8.94 \\
    & YNet & 1.46 & 1.95 & - & 1.34 & 1.43 & - \\
    & Const. Vel. & 2.80 & 11.0 & 25.5 & 2.80 & 11.0 & 25.5 \\
    & \emph{LP$^2$} (Ours) & \textbf{1.22} & \textbf{1.42} & \textbf{1.56} & \textbf{1.06} & \textbf{1.20} & \textbf{1.37} \\
    \bottomrule
\end{tabular}}  
\caption{BoN ADE for different time horizons. $^\star$Denotes privileged methods fine-tuned on hold-out scenes. Best number shown in bold, best non-privileged number in underline.}\label{tab:results_ade}
\vspace{-10px} 
\end{table}

\textbf{2) Time Horizon of Meaningful Prediction.}
A central challenge in long-term human trajectory prediction is the fast growth of possible positions with time.
First, the NLL in Fig.~\ref{fig:results_nll} indicates that our prediction remains relevant for the entire \SI{60}{s}.
Nonetheless, we further analyze this limit by inspecting the progression of probability mass through the CTMC by comparing the distribution $\p(x(t))$ to the steady state distribution $\p(x_{steady})$ that satisfies $Q\cdot\p(x_{steady}) = 0$. 
We find that after \SI{69.9}{\second} in the home and \SI{93.0}{\second} in the office environment, the total variation distance $D_{TV}(\p(x(t)), \p(x_{steady}))$ is below $0.5$, \ie the majority of the predicted probability $\p(x(t))$ is static in the leaf nodes of the interaction sequence tree. 
Interestingly, we experimentally find that the NLL for this quasi-static prediction is still lower than that of the \emph{random goal}.
This indicates that the steady state prediction of our model acts like a static prior about which objects humans are most likely to interact with in general. 
However, although such a prior can be useful in many cases, it no longer reflects a trajectory prediction.
We note that a dynamic prediction beyond this horizon is in principle possible by increasing the depth of the ISP module, however, how long the main modes of the prediction stay meaningful depends on the complexity of the environment and task the human is planing to achieve.

\textbf{3) Importance of the Interaction Duration.}
A key challenge of the task is that a human may interact with entities in the environment for an extended duration, disrupting traditional trajectory prediction methods.
For example, a human might interact with a sink for \SI{15}{s} when washing their hands, but \SI{3}{min} when washing dishes (the action label in both cases may just be ``washing''). 
Having to predict the interaction duration $\tau_i$ adds additional complexity and variance to the LHTP task.
To investigate the impact of this, we study the performance of our method when disregarding the interaction times and predicting only the movement in between interactions.
We find that having to account for the interaction duration increases the NLL on average by $25.5\%$ in the home and $31.7\%$ in the office scene.
Although our ISP module accounts for some of the uncertainty in duration prediction by modeling it as an exponential distribution rather than a single estimated value, this highlights the highly complex nature of our problem on top of estimating the correct interactions.

\textbf{4) Object Instance vs. Semantic Class Prediction.}
To analyze \review{the importance of ISP, we compare our strategy of simplifying the prediction task for the LLM by predicting only the semantic class and reasoning about the instance spatially (Sec.~\ref{sec:approach}), to directly predicting the object instance.}
Tab.~\ref{tab:abl_semantic_instance} shows performance metrics if the \emph{semantic} class or \emph{instance} id is predicted (pred.) or known (GT).
Naturally, knowing the true instance the human will interact with achieves the best performance.
This is most pronounced for $t=30-\SI{60}{s}$, as predicting interactions far ahead is challenging.
Nonetheless, knowing the true semantic class already provides a lot of information and results in only moderately worse performance.

In contrast, when these qualities have to be predicted, we see the inverse trend.
As the LLM performs notably better at predicting only the semantic class compared to the specific instance, the optimality gap is much smaller in the semantic case, indicating  better trajectory prediction. 
To isolate the ISP prediction performance, we also report the top-10 accuracy, \ie
how often the true interaction is among the 10 most likely predictions, in Tab.~\ref{tab:results_interaction_accuracy}.
We observe that predicting the right instance out of $120$ and $107$ objects in the home and office, respectively, is challenging.
In the office, the difference is more pronounced as there are more objects of the same class.

\begin{table}[]
\adjustbox{width=\columnwidth}{%
\begin{tabular}{l|ccc|ccc|ccc}
    \toprule
    Metric & \multicolumn{3}{c}{Avg. NLL [-]} & \multicolumn{3}{c}{Bo5 ADE [m]} & \multicolumn{3}{c}{Bo20 ADE [m]} \\
    \cmidrule(lr){2-4} \cmidrule(lr){5-7} \cmidrule(lr){8-10}
    Timeframe [s]& <10 & <30 & <60 & <10 & <30 & <60 & <10s & <30 & <60 \\
    \midrule
    Semantic-Pred. & \underline{3.60} & \underline{4.86} & \underline{5.72} & \underline{2.51} & \underline{4.87} & \underline{5.25} & \underline{2.28} & \underline{3.71} & \underline{3.54} \\     
    Instance-Pred. & 4.23 & 6.52 & 7.63 & 2.67 & 6.48 & 9.00 & 2.67 & 6.25 & 7.35 \\
    Semantic-GT & 3.11 & 3.60 & 4.21 & 2.65 & 4.51 & 4.44 & 2.37 & 3.45 & 3.15 \\
    Instance-GT & \textbf{2.92} & \textbf{3.14} & \textbf{3.33} & \textbf{2.09} & \textbf{2.98} & \textbf{1.97} & \textbf{2.09} & \textbf{2.98} & \textbf{1.97} \\
    \bottomrule
\end{tabular}}
    
\caption{Performance metrics for class vs. instance prediction. Best number shown bold, best non-GT number underlined.}\label{tab:abl_semantic_instance}
\end{table}

\begin{table}[]
\adjustbox{width=\columnwidth}{%
\begin{tabular}{p{0.1em}l|cc|cc}
    \toprule
    \multirow{2}{0.1em}{\rotatebox[origin=c]{90}{Scene}} & Metric & \multicolumn{2}{c}{Top-10 Instance Acc. [\%]} & \multicolumn{2}{c}{Top-10 Semantic Acc. [\%]} \\
    \cmidrule(lr){3-4} \cmidrule(lr){5-6}
    & Interaction \#& First & Second & First & Second \\
    \midrule
    \multirow{2}{0.1em}{\rotatebox[origin=c]{90}{Office}} 
    & Semantic-Pred. & \textbf{55.6} & \textbf{38.7} & \textbf{71.1} & 81.3 \\
    & Instance-Pred. & 35.5 & 29.0 & \textbf{71.1} & \textbf{83.9} \\
    \midrule
    \multirow{2}{0.1em}{\rotatebox[origin=c]{90}{Home}} 
    & Semantic-Pred. & \textbf{56.7} & 40.0 & \textbf{60.0} & 46.7 \\
    & Instance-Pred. & 43.3 & \textbf{53.3} & 50.0 & \textbf{53.3} \\
    \bottomrule
\end{tabular}}
    
\caption{Top-10 accuracy of the predicted object interactions.}\label{tab:results_interaction_accuracy}
\vspace{-10px}
\end{table}

\textbf{5) Deterministic Prediction vs. Probabilistic CTMC.}    
\review{To isolate the performance of our CTMC-based approach, we compare it to a} deterministic baseline that uses the identical interaction predictions as our method in Tab.~\ref{tab:results_abl_ctmc}. 
Rather than modeling the trajectory evolution probabilistically using a CTMC, it directly infers full trajectories from the auto-regressive prediction and deterministically ``walks'' along these trajectory as time progresses.
This shows that by using the CTMC framework and thereby modeling uncertainties like the human's walking speed and interaction times with the objects in the environment as exponentially distributed probabilistic variables, we can improve our prediction. 

\begin{table}[]
\adjustbox{width=\columnwidth}{%
\begin{tabular}{l|ccc|ccc|ccc}
    \toprule
    Metric & \multicolumn{3}{c}{Avg. NLL [-]} & \multicolumn{3}{c}{Bo5 ADE [m]} & \multicolumn{3}{c}{Bo20 ADE [m]} \\
    \cmidrule(lr){2-4} \cmidrule(lr){5-7} \cmidrule(lr){8-10}
    Timeframe [s]& <10 & <30 & <60 & <10 & <30 & <60 & <10s & <30 & <60 \\
    \midrule
    CTMC & \textbf{3.60} & \textbf{4.86} & \textbf{5.72} & \textbf{2.51} & \textbf{4.87} & \textbf{5.25} & \textbf{2.28} & \textbf{3.71} & \textbf{3.54} \\     
    Deterministic & 4.41 & 6.25 & 6.66 & 2.61 & 4.98 & 5.30 & 2.30 & 3.74 & 3.60 \\
    \bottomrule
\end{tabular}}
    
\caption{Performance metrics for deterministic vs. CTMC-based trajectory prediction.}\label{tab:results_abl_ctmc}
\vspace{-10px}
\end{table}

\section{Limitations}
While our method shows promising performance for long-horizon predictions, it does not yet make optimal use of all available information to refine the early part of the trajectory.
While integrating learning-based methods like \cite{Salzmann20eccv-trajectronpp, mangalam21iccv-ynet} might work well, considering additional input modalities like eye gaze and audio-visual signals could be highly beneficial to improve motion and interaction prediction. \review{To improve long-term reasoning, adding cues about the target agent if it is known (e.g., their habits, their job, etc.) could potentially enhance trajectory predictions. Here, the explicit prompting in our method offers great flexibility as any additional information can simply be fed into the prompt.}

\review{Additionally, as our method uses perfect information about past actions and the environment, exploring mechanisms to deal with imperfect information or noise that may be present is of interest.
Moreover, our method currently relies on single points of interaction in semantically rich human-centric environments. Therefore, it tackles neither human motion without interactions (e.g., moving up and down a corridor), nor interactions that cover a larger area (e.g., sweeping the floor in a room). Expanding our method to account for these cases holds great promise and will be explored in the future.}
Another limitation is that the current approach may not scale well to very high interaction prediction depths, as the tree size grows exponentially and Q is quadratic in the tree size. This could be addressed by more selectively rolling out high-probability interactions \review{or considering fully connected transition matrices if predictions beyond \SI{60}{\second} are desired.
When using LLMs, inference time is also of concern. As we did not optimize for inference speed, trajectory prediction still takes long. However, as several parts of our pipeline can be parallelized and we expect LLMs to become faster, we expect that inference times can be reduced to the order of a few seconds.}
Finally, we currently only consider a single agent. Expanding our approach to multi-agent scenarios presents an exciting opportunity to explore the complex dynamics of multiple interacting agents and the impact of social influences on long-term trajectories. 
We believe the methodology of using \emph{dynamic} scene graphs can extend to moving entities and interactions with them. 
\section{Conclusion}
In this work, we explored the problem of long-term human trajectory prediction of up to \SI{60}{s} \review{in human-centric indoor environments}, including human-object interactions.
We presented \emph{LP$^2$}, a novel approach composed of a LLM-based module to predict human-object interactions and a CTMC-based probabilistic planning method that grounds these interaction sequences into a spatio-temporal distribution over future human positions.
We introduced a novel semi-synthetic dataset of human trajectories in indoor environments annotated with rich human-object interactions and long-horizon trajectories.
We show in a thorough experimental evaluation that our method outperforms prior art in long-term human trajectory prediction methods, especially as the time horizon extends beyond 20\,s and when dealing with trajectories that include complex interactions with the environment. 
We show, that we can predict a meaningful distribution of the human's future location for a horizon up to 60\,s and highlight the importance of modeling the temporal nature of the interactions with the environment.
We will release our code and dataset to facilitate further research in this area.

\bibliographystyle{IEEEtran}
\bibliography{references/refs, references/myRefs}

\end{document}